\documentclass{article}

\usepackage{arxiv}
\usepackage[utf8]{inputenc}
\usepackage[T1]{fontenc}
\usepackage{amsmath}
\usepackage{hyperref}
\usepackage{url}
\usepackage{booktabs}
\usepackage{amsfonts}
\usepackage{microtype}
\usepackage{graphicx}
\usepackage{array}
\usepackage{tabularx}
\usepackage{fvextra}
\usepackage{tcolorbox}
\tcbuselibrary{breakable}
\usepackage{enumitem}
\usepackage{natbib}
\usepackage{longtable}
\usepackage{listings}
\usepackage{xcolor}
\usepackage{rotating}
\graphicspath{{./}}

\title{Automated Big Data Quality Assessment using Knowledge Graph Embeddings}

\author{
Hadi Fadlallah\thanks{Corresponding author. ORCID: 0000-0003-1160-5980}
\\
Saint-Joseph University, Beirut, Lebanon
\\
\texttt{hadi.fadlallah@net.usj.edu.lb}
\and
Rima Kilany\thanks{ORCID: 0000-0002-5710-6901}
\\
Saint-Joseph University, Beirut, Lebanon
\\
\texttt{rima.kilany@usj.edu.lb}
\and
Mitri Haber\thanks{ORCID: 0009-0003-7452-6793}
\\
Saint-Joseph University, Beirut, Lebanon
\\
\texttt{mitri.haber@net.usj.edu.lb}
\and
Ali Jaber\thanks{ORCID: 0000-0002-9180-8384}
\\
Lebanese University, Beirut, Lebanon
\\
\texttt{ali.jaber@ul.edu.lb}
}
\begin{document}

\maketitle

\begin{abstract}
Automated data quality assessment is crucial for managing big data, but existing solutions face challenges in achieving accurate context-aware assessment. This paper presents a novel knowledge-based approach to enhance automated data quality assessment. Our approach utilizes knowledge graph embeddings to predict missing edges between the input dataset's context representation and the relevant quality rules and dimensions within a knowledge graph representing contextual data characteristics and the required quality assessment operations. We surpass conventional practices by integrating diverse representations within the knowledge graph, drawing insights from contextual information from a thorough literature investigation. This integration allows us to develop a comprehensive and context-specific data quality assessment plan tailored to each context. Leveraging the knowledge graph improves our understanding of the input dataset's context, overcoming the limitations of traditional methods that rely solely on strict matching and overlook contextual characteristics. By injecting numerical edge attributes, we assign corresponding weights to each predicted quality measurement, providing a comprehensive data quality assessment plan for the input dataset.

To evaluate our approach, we leverage AmpliGraph, a framework developed and benchmarked by AccentureLabs. The evaluation involves employing a real-world radiation sensors dataset provided by the Lebanese Atomic Energy Commission (LAEC-CNRS). The results obtained from this evaluation demonstrate the capability of our solution to generate a comprehensive data quality assessment plan for the given input dataset.
\end{abstract}

\keywords{Data quality assessment \and Data context \and Big data \and Machine learning \and Knowledge graph embeddings \and Automation}

\section{Introduction}

Data quality assessment is critical in ensuring data reliability, validity, and usability for various data-driven applications \cite{Strong1997DataQI}. It serves as a fundamental step in data management, where different aspects such as data accuracy, completeness, consistency, and timeliness are assessed to make informed decisions and drive meaningful insights \cite{Wang1996BeyondAW}. Manual data quality inspection becomes increasingly challenging when faced with big data due to the wide variety, high velocity, and sheer volume to be examined \cite{Helfert2009ACA}. The human capacity to examine each data set and identify potential errors or inconsistencies, and develop the relevant quality assessment tasks becomes severely constrained. Moreover, this is highly time-consuming, often requiring significant resources and effort. Automated data quality assessment becomes paramount in the era of big data, where data volume, velocity, and variety are continuously growing. Automated approaches leverage computational algorithms, techniques, and tools to assess data quality attributes systematically and efficiently. Organizations can save time, reduce costs, and ensure a more consistent and objective data quality evaluation across large datasets by automating the process \cite{Fadlallah2023ContextAware}.

To address these challenges, we previously introduced BIGQA \cite{Fadlallah2023BIGQA}, a comprehensive framework for automating big data quality assessment. BIGQA fills a gap in current big data quality assessment solutions and makes valuable contributions; the framework introduces a declarative approach that effectively automates data quality assessment operations while minimizing user intervention. By incorporating the ISO/IEC 15939 \cite{ISO15939} and ISO/IEC 25000 \cite{ISO25000} standards, it successfully addresses the unique challenges of big data's volume, variety, and velocity. Additionally, BIGQA presents an innovative architectural design and implementation using prominent big data tools. The solution's architecture aligns with the ISO/IEC 20547 \cite{ISO20547} big data reference architecture, enhancing its robustness and compatibility. Moreover, the framework offers a powerful capability of generating customized data quality reports, providing users with tailored insights. Most importantly, BIGQA has context-aware quality assessment capabilities since it incorporates a context analyzer component that harnesses the characteristics of the data context. It employs machine learning techniques, specifically Node2Vec \cite{grover2016node2vec} and K-Nearest Neighbors \cite{peterson2009k} to retrieve the most relevant data quality assessment plan from a knowledge graph that contains information about data contexts and their relevant quality assessment plans. Still, the context analyzer has limitations in considering the complete range of context characteristics. Moreover, the current implementation of the context analyzer in BIGQA focuses solely on selecting the quality assessment plan for the most similar data context in the knowledge graph. Focusing only on the most similar context could result in losing important details required to assess the incoming dataset accurately. The knowledge graph may also contain other data contexts providing additional information that could help create a more accurate and thorough assessment approach for the input dataset.

\begin{figure}[htbp]
\centering
\centerline{\includegraphics[width=\columnwidth]{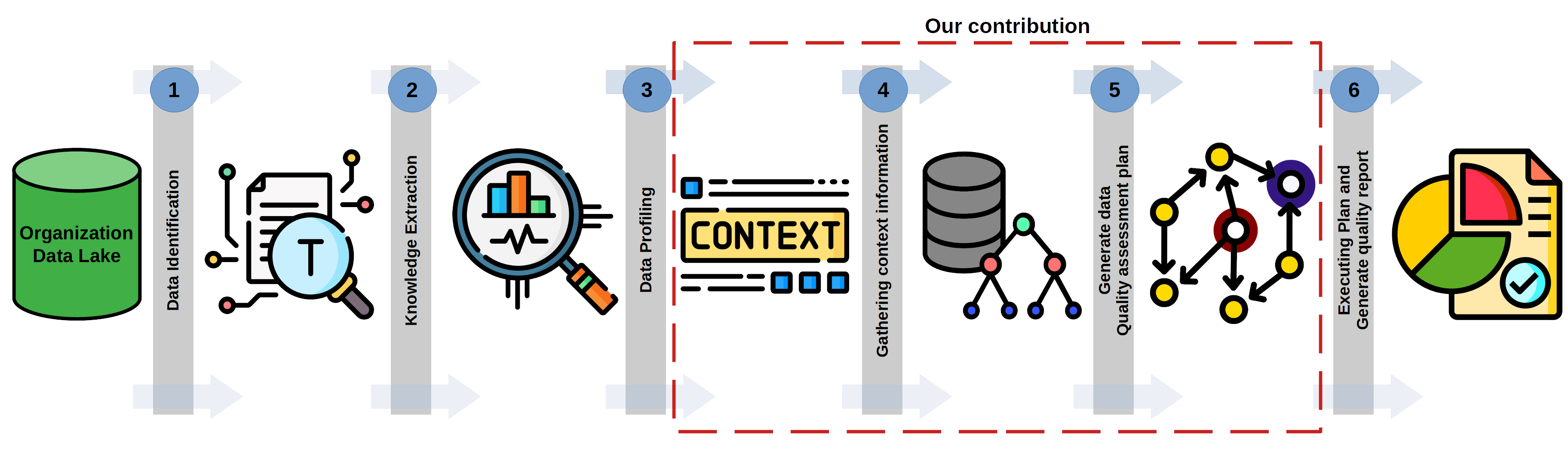}}
\caption{Gathering context information and generating a data quality assessment plan}
\label{fig:flow}
\end{figure}

To overcome these limitations, this paper focuses on enhancing the steps of gathering context information and generating a tailored data quality assessment plan to ensure accurate and context-specific execution (Figure~\ref{fig:flow}); once a plan is generated, it will be converted to an executable code that will generate as expected the needed data quality report. This paper begins by investigating the literature to identify a comprehensive description of data context characteristics and develop the most complete and accurate representation of the data context. Data contexts and their attributes, including data size, structure, attribute types, domain, and file formats, are described in the knowledge graph. A dedicated data quality assessment plan is integrated and linked with the corresponding context characteristics by data experts or researchers. Each plan comprises several quality rules (base measures) and dimensions (derived measures). When dealing with a new data set, the context information is extracted and added to the knowledge graph; however, the link to a valid data quality assessment plan does not exist. This paper presents an enhanced approach that leverages knowledge graph embedding to predict missing edges between the input data context representation and the existing quality rules and dimensions within the knowledge graph. Our proposed method involves numerical edge injection attributes into knowledge graph embeddings to add weights to the predicted quality measurements, resulting in a complete data quality assessment plan.

Injecting numeric edge attributes into knowledge graph embeddings is essential as it allows the model to capture additional information related to uncertainty, edge importance, and out-of-band knowledge. This enhancement improves the model's predictive capabilities and enables it to generate more accurate and context-specific data quality assessment plans. Integrating literature-derived context characteristics into our methodology enables us to overcome potential disparities between the input dataset and the retrieved quality assessment plan's schema, resulting in improved accuracy and effectiveness of the assessment process.

We evaluated our solution using a Neo4j\footnote{https://neo4j.com} knowledge graph representing the data context characteristics and the related data quality assessments alongside a knowledge graph embedding solution called AmpliGraph \cite{luca_costabello_2023_7709418} developed by Accenture Labs. The results demonstrate that our approach provides a more accurate data quality assessment plan compared to the one retrieved by the BIGQA context analyzer component. The evaluation was conducted on a real-world dataset from radiation sensors provided by the Lebanese Atomic Energy Commission (LAEC-CNRS)\footnote{http://www.laec-cnrs.gov.lb/}.

The remainder of this paper is organized as follows. Section 2 presents an overview of the fundamental concepts that form the basis of this research. Section 3 provides a literature review on knowledge graph embedding and existing automated data quality assessment approaches. Section 4 describes the proposed solution. Section 5 presents the experiment and evaluation, including a description of the datasets used and a comparison of the proposed method with BIGQA. Section 6 concludes the paper and presents future research directions.

\section{Background}

This section overviews knowledge graph embeddings and briefly explains our proposed solution for automating data quality assessment.

\subsection*{Knowledge graph embeddings} 
\label{KGE}
Graph machine learning models have emerged as powerful tools for solving complex problems and tasks that are not easily represented in tabular form. Three main families of graph machine learning models have been established: node representation models, graph neural networks, and knowledge graph embeddings. Each family comprises multiple models that have successfully addressed specific problem domains. A comprehensive survey by Makarov et al. \cite{makarov2021survey} provides an extensive overview of graph machine learning models and their various use cases.

Knowledge graph embeddings are a class of graph representation learning methods that learn vector representations of the nodes and edges of a graph. They are applied to graph completion, knowledge discovery, entity resolution, and link-based clustering. The goal is to learn a low-dimensional vector representation for each entity and relation in the graph, such that the pairwise similarity between vectors reflects the semantic similarity between the corresponding entities and relations. The main limitation of traditional knowledge graph embeddings is that they are not designed to capture numeric values associated with edges, which can represent uncertainty, edge importance, and out-of-band knowledge in many scenarios \cite{dai2020survey}.

Pai et al. \cite{pai2021learning} proposed a novel method called \textit{FocusE} that injects numeric edge attributes into the scoring layer of a traditional knowledge graph embedding architecture. This method considers numeric literals associated with edges and treats triples with low-numeric values as pseudo-negatives.

\begin{figure}[htbp]
\centering
\centerline{\includegraphics[width=\columnwidth]{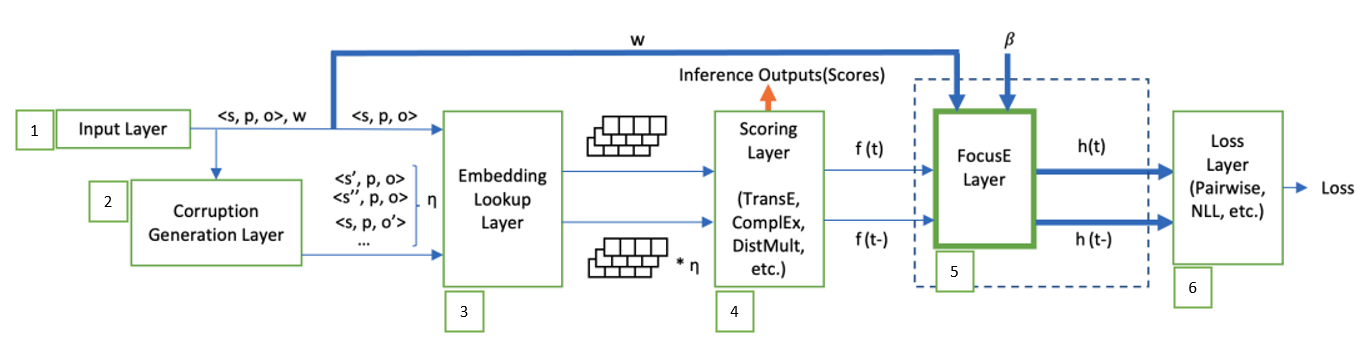}}
\caption{A knowledge graph embedding model architecture enhanced with FocusE \cite{pai2021learning}}
\label{fig:pai2021learning}
\end{figure}

The architecture of the proposed \textit{FocusE} model is illustrated in Figure \ref{fig:pai2021learning}. Here is an overview of how the solution works:

\begin{enumerate}
\item The input of the model is a list of sets, where each set represents a relation between two nodes and includes the subject, predicate, object, and weight $(s, p, o, w)$.
\item A corruption layer generates incorrect or corrupt triples for each input triple to enhance the model's ability to distinguish true triples from their corruptions.
\item An embedding layer, represented by a \(k \times n\) matrix and a lookup function, maps nodes to low-dimensional vectors, where $k$ is the width or length of a vector that represents a node (a hyperparameter to be optimized), and $n$ is the number of nodes in the graph.
\item A scoring layer is applied to the subject, predicate, and object vectors from the embedding matrix. This layer must be expressive enough to differentiate between different relations and identify incorrect ones. In the case of \textit{FocusE}, the \textit{ComplEx} function is utilized as the scoring layer, which calculates the real term of the Hermitian dot product of the vectors in the complex space \cite{trouillon2016complex}.
\item FocusE modulates the output of the scoring layer based on the associated triplet $<s, p, o>$. It helps the model focus on triples with higher numeric values during training, thereby improving its ability to distinguish true triples from their corruptions.
\item The output of \textit{FocusE} is used in the loss function of the knowledge graph embeddings model.
\end{enumerate}

By incorporating the \textit{FocusE} method into the knowledge graph embedding architecture, our solution can effectively predict missing edges and then capture and utilize numeric information associated with graph edges, enhancing the model's predictive capabilities and improving the quality assessment plan for new datasets.

\section{Literature Review}

This section examines the representation of data context characteristics and explores existing approaches for automating data quality operations in data-driven applications.

\subsection{Data Context}
\label{ctx}

Making more accurate decisions requires assessing data quality levels. Different factors must be considered to assess data quality, especially the data context. Due to the contextual diversity of real-world use cases, different entities, such as individuals and organizations, consider the data context as a critical factor. The data used in one context (e.g., an organization’s policy) may not be effective in another. As decision-making relies heavily on data, measuring data quality based on its intended use is essential. In the literature on data quality assessment, various data context definitions are used to address the different aspects of data quality \cite{Fadlallah2023ContextAware}: 

\begin{itemize}
    \item Customer perspective, as defined by Even et al. \cite{even2005value} \cite{even2007utility}.
    \item The domain of the data, as emphasized by Taleb et al. \cite{taleb2015big} \cite{taleb2018big} and Cai et al. \cite{cai2015challenges}.
    \item Application scope, which considers the specific applications and use cases, as noted by Merino et al. \cite{merino2016data}.
    \item Organizational and decision-making policies, highlighting the influence of internal policies, according to Immonen et al. \cite{immonen2015evaluating} and Paakkonen et al. \cite{paakkonen2015reference}.
    \item Data source type, focusing on the characteristics and origin of the data, as indicated by Batini et al. \cite{Batini2011ADQ}.
    \item The data quality task and the information system environment, considering the tasks and environment in which data is processed, as discussed by Helfert et al. \cite{helfert2009context} and Bronselaer et al. \cite{bronselaer2018operational}.
\end{itemize}

Based on the literature investigation, we identified that data context elements could be categorized into different types (Figure~\ref{fig:Context}):

\begin{figure}[htbp]
\centering
\centerline{\includegraphics[width=\columnwidth]{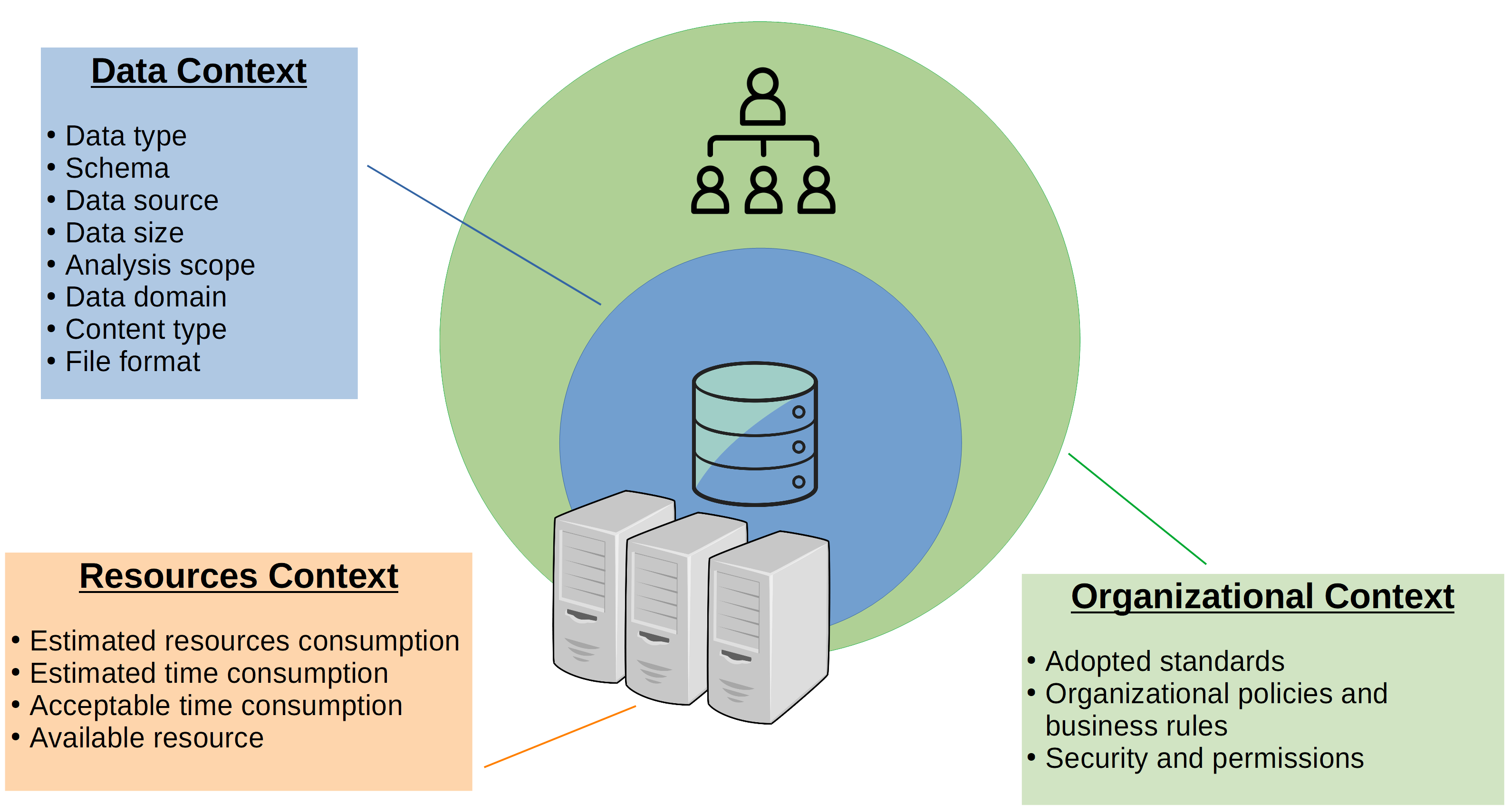}}
\caption{Data context characteristics}
\label{fig:Context}
\end{figure}

\begin{itemize}
\item \textbf{Data context}:
\begin{itemize}
\item Data type: This refers to the structure of the data, which can be classified as structured, semi-structured, or unstructured.
\item Schema: Describes the entities, attributes, and attribute types present in the data.
\item Data source: Represents the origin of the data, which can include machine logs, sensor feeds, social media posts (tweets), short message service (SMS), and more.
\item Data size: Indicates the volume or size of the data.
\item Analysis scope: Specifies the time range or data scope for analysis.
\item Data domain: Refers to the specific domain or field to which the data belongs, such as social media, Internet of Things (IoT), or news feed.
\item Content type: Describes the nature or topic of the data, such as fun, sports, politics, etc.
\item File format: Specifies the format in which the data is stored, such as JSON, XML, or relational database (RDB).
\end{itemize}
\item \textbf{Organizational context}:
\begin{itemize}
    \item Adopted standards: Identifies any standards or guidelines the organization follows in managing and processing the data.
    \item Organizational policies and business rules: Refers to the specific policies and rules established by the organization governing data management and decision-making processes.
    \item Security and permissions: Encompasses the security measures and access permissions associated with the data, ensuring data confidentiality and integrity.
\end{itemize}

\item \textbf{Resources context}:
\begin{itemize}
    \item Estimated/available resources consumption: Indicates the estimated or available resources required to process and analyze the data, including computational power, storage, memory, etc.
    \item Estimated/available time: Specifies the estimated or available time for performing data-related tasks, such as data cleaning, analysis, and decision-making.
\end{itemize}
\end{itemize}

These distinct data context elements provide a comprehensive framework for understanding the contextual factors influencing data quality and decision-making processes. By considering these elements, organizations can gain a deeper understanding of their data and make more informed decisions based on the specific context in which the data is utilized. The context information can be extracted from the data using various techniques:

\begin{itemize}
\item \textbf{Knowledge extraction}: Named-Entity Recognition can extract domain-specific terms, date values, and other relevant information.
\item \textbf{User specification}: End-users can explicitly specify certain aspects of the context, such as time scope, policies, or desired response time.
\item \textbf{Calculation}: Contextual factors like estimated time and resource consumption can be calculated based on available data and system characteristics.
\end{itemize}

By leveraging these techniques, organizations can effectively capture and utilize the context information inherent in their data during data quality assessment or further analysis.

\subsection{Automating data quality operations}

The literature presents various automated solutions for data quality operations. Kedad et al. \cite{Kedad2002OntologyBasedDC} utilized domain knowledge ontologies to address value inconsistencies between different sources, aiming to improve data cleaning processes. Wang et al. \cite{Wang2005AnOA} introduced a data cleaning ontology that stored data cleaning tasks, methods, and reasons, treating data cleaning as a systematic procedure. Wei-Liang et al. \cite{Weiliang2009AnchoringTC} proposed a solution incorporating a central ontology defining unified schema and business rules. These rules were utilized during the data integration process to address conflicts related to schema definitions, value inconsistencies, and inconsistent rule violations across multiple data sources. Oliveira et al. \cite{Oliveira2006AnOA} developed a data cleaning knowledge base using a domain ontology specifically for checking and repairing data completeness, accuracy, and consistency. However, mapping the ontology and the database schema was performed manually. Missier et al. \cite{Missier2006QualityVC} proposed a framework for expressing and applying quality-based, personal data acceptability criteria, called quality views, in data processing environments used by life scientists. The framework included a user-extensible semantic model for information quality concepts, a process model, a simple declarative language for specifying abstract quality views, and an architecture for implementing and deploying quality views within various data processing environments. Br{"u}ggemann et al. \cite{Brggemann2009UsingOP} presented three applications of ontologies in data quality management: context-aware consistency, duplicate detection, and metadata annotation. Choi et al. \cite{Choi2008AnEM} proposed a data quality service that utilized a user-developed ontology to validate data based on defined semantic and business rules. An et al. \cite{An2006BuildingSM} proposed a solution for discovering possible relationships between the data source schema (relational or XML) and an ontology.  Milano et al. \cite{Milano2005UsingOF} introduced the OCX framework, which leveraged a domain knowledge ontology for XML data cleaning.

Ehrlinger et al. \cite{ehrlinger2017automated} introduced a high-level architecture for a data quality assessment solution to assess and monitor data quality across heterogeneous data sources. This architecture comprises four components: (1) a data profiling component, (2) a data quality repository, (3) time-series analytics, and (4) a user interface. At the core of this solution lies the data quality repository, which includes an ontological description of the assessed information system schema and a database specifically designed to store data quality metrics, including those relevant to time-series analysis. Time-series analytics is crucial in detecting temporal data outliers that can significantly impact data quality. By incorporating this analysis, the solution provides a mechanism to identify and address data inconsistencies over time. While this proposal offers a comprehensive high-level architecture, certain aspects, such as integrating data sources with the ontological description in the quality repository, may require further elaboration. Nonetheless, using such a repository proves highly beneficial in addressing data variety by establishing linkages between data elements through a unified description.

Furthermore, several research studies have focused on constructing and demonstrating domain knowledge ontologies tailored to specific domains. For instance, Frank \cite{Frank2007DataQO} showcased using a domain knowledge ontology to assess GIS data quality, while Liaw et al. \cite{Liaw2013TowardsAO} conducted a review to develop a domain knowledge ontology for data quality in the integrated chronic disease management domain. Johnson et al. \cite{johnson2015data} proposed a data quality ontology for electronic health records (EHR) data, and Urrutia et al. \cite{Urrutia2017AnOT} demonstrated a data quality ontology for the healthcare domain. Geisler et al. \cite{Geisler2011OntologybasedDQ} introduced an ontology-based data quality solution specifically designed for Data Stream Management Systems. However, these proposals have limitations, such as relying on exact matching techniques to map the input dataset schema to the knowledge graph and overlooking crucial data context characteristics.

\begin{figure}[htbp]
\centering
\centerline{\includegraphics[width=\columnwidth]{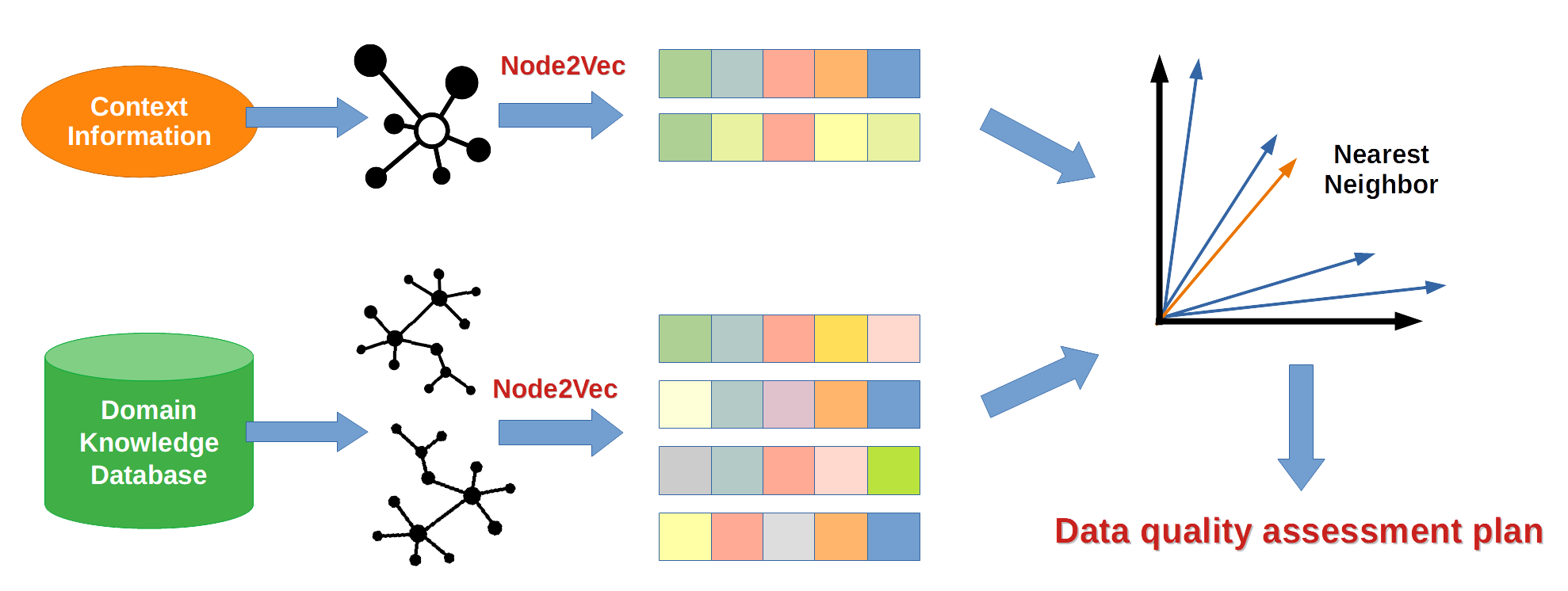}}
\caption{BIGQA context analyzer workflow \cite{Fadlallah2023BIGQA}}
\label{fig:Context2Vec}
\end{figure}

In a previous paper, we presented BIGQA \cite{Fadlallah2023BIGQA}, a data quality assessment framework with a component dedicated to automating quality assessment operations and reducing the reliance on user intervention. We proposed a context analyzer component to enhance the solution's adaptability across diverse data contexts. This component utilizes a knowledge graph to enable context-driven data quality assessment by capturing metadata about data sources and their relationships (see Figure~\ref{fig:Context2Vec}). The knowledge graph is generated and maintained based on the organization's data environment, encompassing relevant data characteristics. The knowledge graph contains information about data contexts and their characteristics, such as data size, schema, attribute types, domain, and file formats. Moreover, each context has a corresponding data quality assessment plan composed of several quality rules (base measures) and dimensions (derived measures).

The context analyzer automates data quality assessment by leveraging data context attributes and employing machine learning techniques, specifically Node2Vec \cite{grover2016node2vec} and K-NN \cite{peterson2009k}. The K-NN algorithm is applied to compare the context of the input data with existing contexts in the knowledge graph. The corresponding data quality assessment plan is retrieved if a similar context is identified based on a preconfigured threshold.
By incorporating the K-NN algorithm and establishing a similarity threshold, we ensure that the retrieved data quality assessment plan aligns closely with the context of the input data. This automated approach reduces the reliance on user intervention and enables efficient and accurate data quality assessment.

One primary limitation of this solution is its exclusive reliance on the quality assessment plan derived from the most relevant data context within the knowledge graph. However, this approach may not always provide the necessary information tailored to the input dataset. Other data contexts within the knowledge graph may contain valuable insights and details that could contribute to a more accurate and precise assessment plan. By solely relying on the most pertinent data context, there is a risk of overlooking potentially valuable information that could enhance the quality assessment process. Therefore, it is essential to consider the broader range of data contexts within the knowledge graph to ensure a comprehensive assessment plan.

\section{Automated Knowledge-Driven Data Quality Assessment}

This section presents our proposed automated data quality assessment solution, which consists of two main subsections. The first subsection focuses on constructing a data context knowledge graph, detailing the representation and organization of context characteristics and quality assessment plans. The second subsection introduces our novel automated data quality assessment approach using knowledge graph embeddings.

\subsection{Data context knowledge graph}

A knowledge graph is a powerful representation of structured and interconnected knowledge, where entities are represented as nodes and relationships between them are depicted as edges \cite{StanfordKG}. In the context of automating tasks, a knowledge graph serves as a crucial component, enabling efficient data management, analysis, and decision-making.

In our proposed solution, the data representation in the knowledge graph is designed to capture the diverse characteristics of each data context. Each data context, encompassing attributes defined in section~\ref{ctx}, is represented as a node within the knowledge graph. Furthermore, each context is related to a schema node linked to groups of attributes, with each attribute having a specified type (date, text, numeric), forming a structured hierarchy.

The knowledge graph also incorporates quality rules and dimensions as nodes. Quality rules are associated with specific attributes, reflecting the criteria for assessing data quality. Quality dimensions are linked to one or more quality rules, providing a comprehensive framework for evaluating different aspects of data quality. The edges in the knowledge graph are weighted to capture the importance and relevance of different relationships. This weighting scheme allows for a nuanced representation of the connections between attributes, quality rules, and dimensions, facilitating a more accurate data quality assessment.

\begin{figure}[htbp]
\centering
\centerline{\includegraphics[width=\columnwidth]{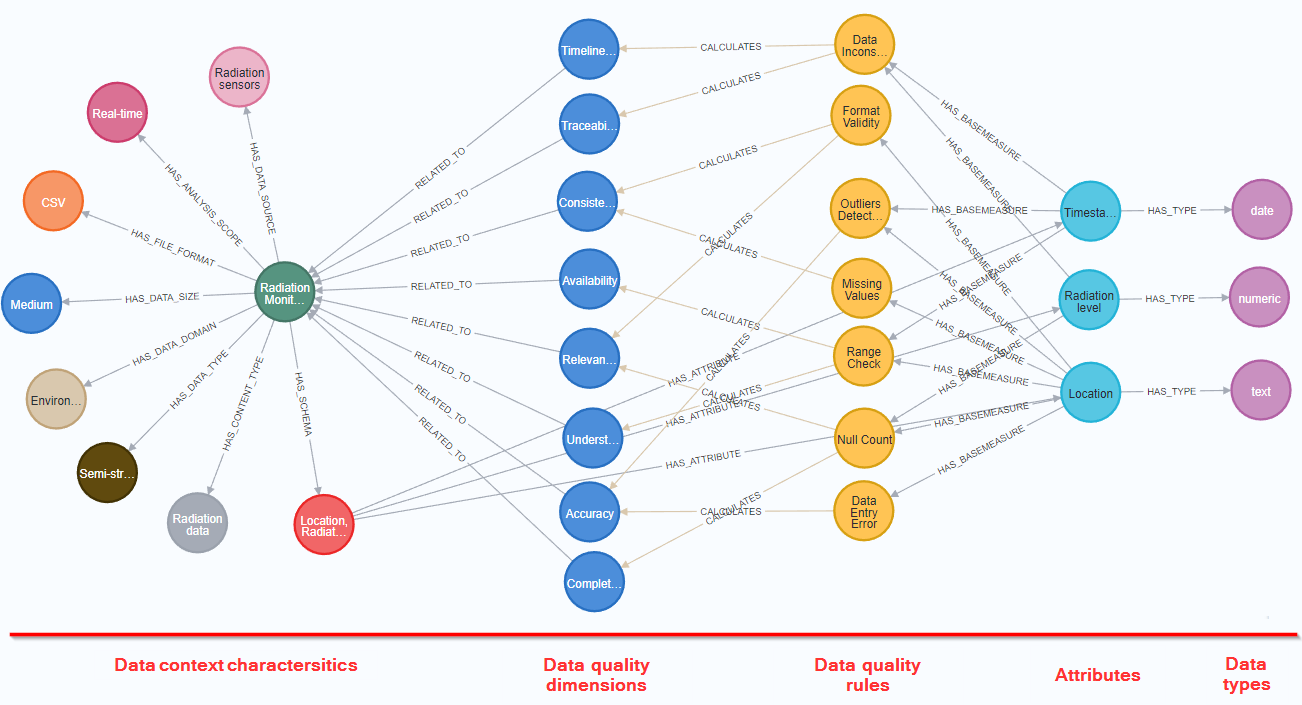}}
\caption{A representation of a data context with the related data quality assessment}
\label{fig:KGExample}
\end{figure}

A data quality assessment plan is a sequence of operations represented as nodes. It represents the desired data quality assessment operations for a specific data context. The plan mainly contains the quality rules operations (base measures), such as the null values ratio and outliers ratio, and the quality dimensions (derived measures) calculated from those measures, such as data completeness and accuracy. The knowledge graph represents a quality assessment plan specific to each data context as a set of edges. These edges connect the attributes of a particular context to a group of quality rules and dimensions. Each edge is weighted with a percentage value. This structured representation (Figure~\ref{fig:KGExample}) is the primary building block we rely on to automate the data quality assessment by traversing the graph and evaluating each context's defined rules and dimensions.

\subsection{Proposed solution}

In the context of predicting a quality assessment plan for a new dataset, the role of a graph ML model is to predict the links and weights between the data schema attributes, the quality rules, and the quality dimensions that exist in the knowledge graph. Numeric weights have proven valuable in expressing uncertainty, edge significance, and supplementary knowledge in various domains, such as genetic data and social networks within knowledge graphs. However, traditional models for embedding knowledge graphs often struggle to incorporate this type of information, which limits their predictive capabilities.

To address this limitation, we employed the novel knowledge graph embedding model (Section~\ref{KGE}) benchmarked by Accenture labs \cite{pai2021learning}. This model introduces two key improvements to enable weight prediction between nodes:

\begin{itemize}
\item Injection of numeric edge attributes into a traditional knowledge graph embedding architecture scoring layer, allowing the model to capture and utilize numeric information associated with graph edges.
\item Open World Assumption (OWA): Embracing the notion that the absence of a fact or a relation between nodes does not imply its falsehood but indicates that the information is unknown. This assumption enhances the model's ability to reason and predict uncertain scenarios with incomplete information.
\end{itemize}

\begin{figure}[htbp]
\centering
\centerline{\includegraphics[width=\columnwidth]{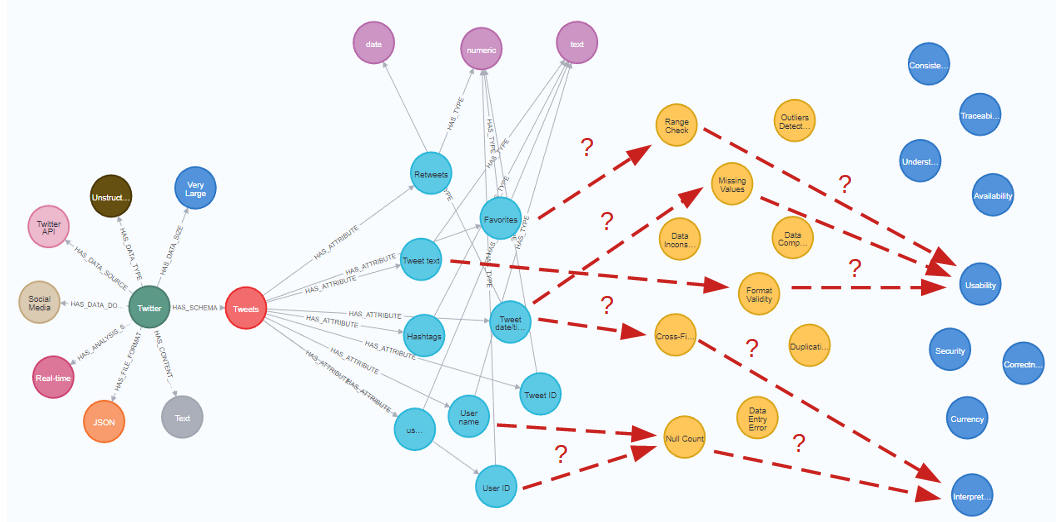}}
\caption{Predicting possible data quality assessment plan of a new data context}
\label{fig:missing}
\end{figure}

By incorporating these enhancements, our knowledge graph embedding model can effectively predict weights and capture nuanced relationships between nodes, enabling the prediction of a relevant data quality assessment plan for an input dataset (Figure~\ref{fig:missing}). The edges and weights we are particularly interested in predicting are those associated with quality rules and dimensions, as they constitute the main components of a comprehensive quality assessment plan. Leveraging our model, we can accurately estimate the weight reflecting the importance and impact of each quality rule and dimension, enabling an informed evaluation of data quality. This predictive capability enhances the utility of our solution in automating the generation of tailored quality assessment plans.

\begin{figure}[htbp]
\centering
\centerline{\includegraphics[width=\columnwidth]{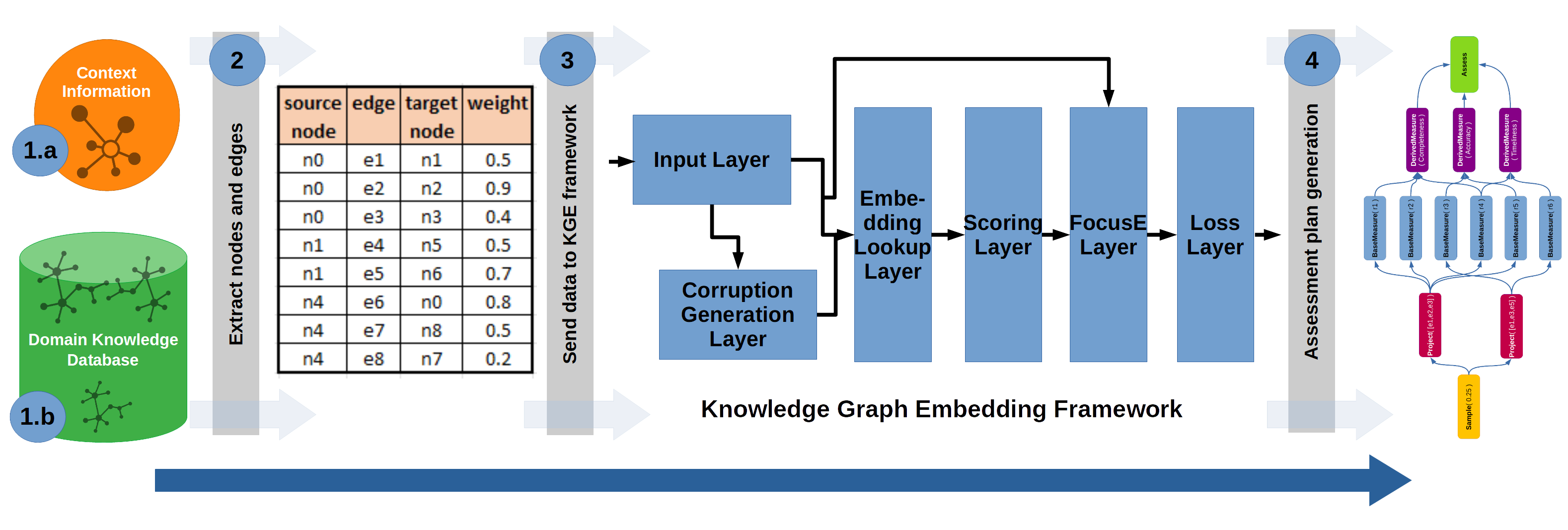}}
\caption{Generating data quality assessment plan using a knowledge graph embedding model}
\label{fig:KGE_QA}
\end{figure}

Our solution works as follows (Figure~\ref{fig:KGE_QA}):
\begin{enumerate}
\item Read the existing data contexts' characteristics
\begin{enumerate}
    \item Input dataset context: the flow begins with extracting context characteristics from the input dataset.
    \item Read existing contexts and their quality assessment plans from the knowledge graph.
\end{enumerate}
\item Convert the extracted context characteristics and the existing context and quality assessment plan into a list of sets (source node, target node, edge, weight).
\item Pass the generated list of sets as input to the knowledge graph embeddings solution where the steps we have already detailed in section~\ref{KGE} are applied.
\item Utilize the knowledge graph embedding model to manipulate the data and generate the quality assessment plan for the input dataset.
\end{enumerate}

The knowledge graph embedding model comprehensively understands the dataset's structure and characteristics by incorporating the input dataset's schema, attributes, data types, domain, file types, and data size. Combined with the existing knowledge graph data and context, this enriched information allows the knowledge graph embedding model to capture intricate relationships, semantic similarities, and contextual dependencies.

Through its advanced learning algorithms, the knowledge graph embedding model leverages this integrated knowledge to generate all the necessary information for the quality assessment plan of the input dataset. This includes predicting links, weights, and correlations between different nodes within the knowledge graph. The result is a complete quality assessment plan considering the input dataset's specific context and characteristics, enabling organizations to automate quality assessment operations and make informed decisions.

\section{Evaluation}

This section presents a comprehensive assessment of the effectiveness of both approaches: the BIGQA context analyzer and the proposed enhanced solution. It highlights the differences between their capabilities in generating an accurate data quality assessment plan of a real-world use case dataset.

\subsection{Dataset}

As an input dataset, we used the radiation sensors data set provided by the Lebanese Atomic Energy Commission (LAEC-CNRS) \cite{Fadlallah2018RaDEnAS} \cite{Fadlallah2019ORADIEXAB}. The radiation sensors dataset is a complex dataset consisting of diverse attributes, including radiation dose rate (numeric), location (text), battery level (numeric), rain level (numeric), and timestamp (date).

\subsection{Knowledge Graph}

We utilized a Neo4j graph database\footnote{https://neo4j.com} to construct a knowledge graph encompassing contextual characteristics of data contexts. Initially, we manually created a data context graph incorporating the characteristics outlined in section~\ref{ctx}. Subsequently, we employed the ChatGPT\footnote{https://chat.openai.com/} machine learning model to generate similar data for an additional forty data contexts.

We first established nodes for all data quality dimensions specified in the ISO/IEC 25012 \cite{ISO25012-1} standard to develop the data quality assessment plan within these contexts. Additionally, we included ten nodes representing various data quality measures; missing values, data inconsistency, null count, data entry error, outliers detection, data comparison, cross-field validation, range check, duplication check, and format validity. Finally, we programmatically generated a randomized data quality assessment plan for each context. 

Consequently, the knowledge graph encompasses forty-one data contexts, each characterized by multiple attributes and a quality assessment plan comprising various quality dimensions and measures (Figure~\ref{fig:KGExample}). We will utilize a comprehensive list of sets extracted from the knowledge graph throughout the experiments. This list encompasses a total of 5877 sets, consisting of source nodes, target nodes, edges, and weights.

\subsection{Experiments}

We conducted two experiments: 
\begin{enumerate}
    \item Generating a data quality assessment using the BIGQA context analyzer by applying Node2Vec and K-NN.
    \item Generating a data quality assessment utilizing the AmpliGraph \cite{luca_costabello_2023_7709418}.
\end{enumerate}

\subsubsection{\textbf{Implementing BIGQA context analyzer}} \hfill

We developed a Python script to implement the BIGQA context analyzer for data quality assessment. This script executes the entire flow, which includes retrieving data context characteristics from the knowledge graph, converting them into a low-dimensional vector representation, and identifying the most relevant data quality assessment plan based on the input dataset's context. 

\subsubsection*{\textbf{The workflow}} \hfill

The script executes the following steps:

\begin{enumerate}
  \item Extract the current data context characteristics from the input dataset and the existing context from the knowledge graph. Store them as nodes and relationships in a CSV file.
  \item Take the generated CSV file as input and convert it into an adjacency matrix.
  \item Create and train a Node2Vec model based on the adjacency matrix.
  \item Apply the cosine similarity of the K Nearest Neighbors (K-NN) algorithm using the Node2Vec model to identify the most similar context to the current data context.
  \item Return the corresponding data context ID.
  \item Input a data context ID and retrieve the associated data quality assessment plan from the knowledge graph.
\end{enumerate}

\subsubsection*{\textbf{Retrieving a data quality assessment plan for radiation data}} \hfill

\begin{figure}[htbp]
\centering
\centerline{\includegraphics[width=\columnwidth]{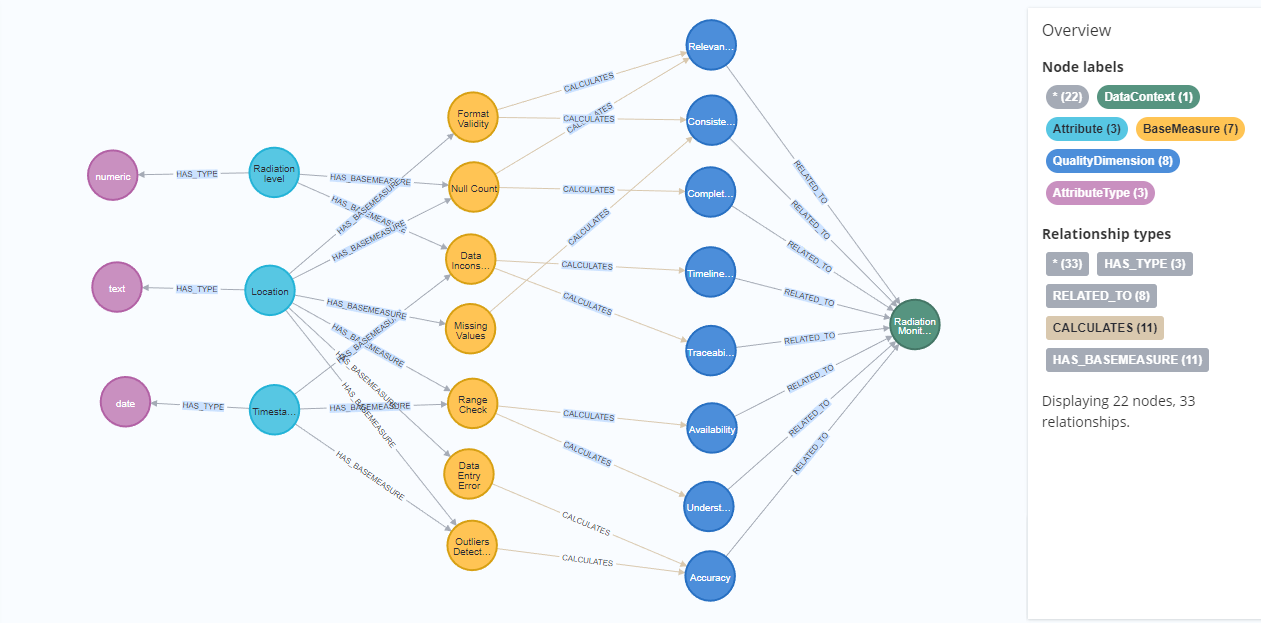}}
\caption{Retrieved data quality assessment plan using BIGQA context analyzer}
\label{fig:Plan}
\end{figure}

The results were promising, as a data quality assessment plan related to a similar data context was successfully retrieved from the knowledge graph. The plan is related to a structured radiation monitoring dataset already defined in the knowledge graph. This plan included seven quality rules measured on three attributes: radiation level, sensor location, and measurement time. These seven quality rules were used to calculate eight quality dimensions (Figure~\ref{fig:Plan}). It is worth noting that while these attributes were considered in the assessment plan, two attributes in the input dataset should have been considered: battery level and sensor ID. Nonetheless, the retrieved assessment plan provided valuable insights into the quality of the data and helped identify potential areas for improvement.

\subsubsection{\textbf{Implementation using AmpliGraph}} \hfill

Our solution was implemented using AmpliGraph, a Python library for representation learning in knowledge graphs; Accenture built it on top of TensorFlow \cite{abadi2016tensorflow}.

Like machine learning models, the knowledge graph model described in section~\ref{KGE} has hyperparameters that need to be explored and set for a specific use case or graph type before automating a pipeline for training the model on incoming datasets.

To determine the optimal hyperparameters, we conducted a grid search with multiple combinations using the grid search feature provided by AmpliGraph. The grid search process involved exploring various combinations of hyperparameters. Each combination was trained for 2000 epochs using the Adam optimizer. The grid search process took approximately 60 hours of computation using a V100 machine on Google Colab\footnote{https://colab.research.google.com/}. Ultimately, we selected the combination of hyperparameters that resulted in the lowest test loss. These optimal parameters are used in new trainings which will take approximately 30 minutes.

For the loss function, we employed the pairwise loss function, which aggregates over the examples and their generated corrupted triplets. This loss function aims to penalize cases where the scoring function of the positive example is smaller than the negative example's scoring function plus a margin gamma. The pairwise margin-based hinge loss equation is given by:

\[
L_{\text{pairwise}} = \sum_{t_+} \sum_{t_-} \max(0, (\gamma + f_{t_+}) - f_{t_-})
\]

Where \(t_+\) denotes a positive example, \(t_-\) denotes a negative example, and \(f_{t_+}\) and \(f_{t_-}\) represent the scoring functions of the positive and negative examples, respectively. 

\subsubsection*{\textbf{Parameters}} \hfill

Below are the parameters determined based on the grid search we conducted:

\begin{table}[htbp]
    \centering
    
    \begin{tabular}{c c}
        \hline
        \textbf{Parameter} & \textbf{Value} \\
        \hline
        Optimizer & Adam\\
        Negative corruption entities & All \\
        Regularizer & L4 \\
        Batch size & 64 \\
        K & 50 \\
        ETA & 5 \\
        Optimizer learning rate & 1e-05 \\
        Loss Function & Pairwise \\
        Loss margin & 0.5 \\
        \hline
    \end{tabular}
    \caption{Optimal parameters}
\end{table}

\begin{enumerate}
  \item \textbf{Optimizer}: The optimization algorithm used during training. Adam is a popular optimizer that combines adaptive learning rates with momentum.
  \item \textbf{Negative corruption entities}: The limitation for generating negative corrupted triplets during training. \textit{"All"} indicates that all entities from the training dataset are considered for corruption. In contrast, \textit{"Batch"} indicates that only entities from the batch can be considered to generate corrupted triplets. Mainly two benefits of choosing the \textit{"All"} method: firstly, this method is more efficient because it does not need to consider the entities in the batch. Secondly, it can cover a wider range of entities, which can help the model learn to distinguish between correct and incorrect statements.
  \item \textbf{Regularizer}: L regularization of norm 4 (L4) was used in the optimization process. L4 regularization serves multiple purposes in improving the performance of a model. Firstly, it reduces overfitting by constraining the model's weights from growing excessively. This constraint helps prevent the model from becoming too specialized and ensures it can generalize well to unseen data. Additionally, L4 regularization enhances the stability of the model by mitigating its sensitivity to the training data, leading to more consistent and reliable predictions.
  \item \textbf{Batch size}: The batch size parameter refers to the number of training examples or samples processed in a single forward and backward pass during model training. It is a crucial parameter that compromises the model's training time and generalization.
  \item \textbf{K}: The embedding vector dimension. Each node and relation have a representation vector in the embedding space. Choosing a small value for the vector size may result in an under-representation of nodes and relations, while choosing a large vector may waste GPU memory and over-represent the nodes.
  \item \textbf{ETA}: In the optimization process, several fake triples are generated and compared to each real one. ETA is a parameter that determines the number of fake triplets generated for each true triplet.
  \item \textbf{Optimizer learning rate}: The learning rate controls the step size at each iteration during the optimization process. The learning rate value guides the optimization algorithm in deciding the magnitude of updates to the model's parameters in the backpropagation phase, including embeddings. The value of the learning rate was picked by trial and error. 
   \item \textbf{Loss Function}: Pairwise loss function was used to penalize the discrepancy between the scores of fake triplets and true ones. It tells the model that the scoring of each true example should be higher than a fake one by a margin, or else the loss function will score a penalty.
  \item \textbf{Loss margin}: The loss margin used in the pairwise loss function.
\end{enumerate}

\subsubsection*{\textbf{Performance measurements}} \hfill

The model was re-initialized and retrained using the above parameters for 5000 epochs. The list of sets (source node, target node, edge, weight), which consists of 5877 sets, extracted from the knowledge graph, was divided into training (80\%) and testing (20\%) sets. Below are the performance metrics obtained from cross-validating the model on the test data.

\begin{table}[htbp]
    \centering
    \begin{tabular}{c c}
        \hline
        \textbf{Measure} & \textbf{Value} \\
        \hline
        Loss & 9.45 \\
        MRR & 0.19 \\
        MR & 38.99 \\
        Hits@10 & 0.66 \\
        Hits@3 & 0.26 \\
        Hits@1 & 0.21 \\
        \hline
    \end{tabular}
    \caption{Performance measurements}
\end{table}

\begin{itemize}
    \item \textbf{Loss}: The loss function's value quantifies the discrepancy between the predicted and true values. In this case, the loss value is 9.45.
    \item \textbf{MRR (Mean Reciprocal Rank)}: It measures the average reciprocal rank of the correctly predicted item. A higher MRR value indicates better performance. Here, the MRR value is 0.19.
    \item \textbf{MR (Mean Rank)}: It represents the average rank of the correctly predicted item. Lower MR values indicate better performance. In this case, the MR value is 38.99.
    \item \textbf{Hits@10}: It measures the proportion of test cases where the correct answer is ranked within the top 10 predictions. The Hits@10 value is 0.66, indicating that the correct answer appears in the top 10 predictions in 66\% of the cases.
    \item \textbf{Hits@3}: It measures the proportion of test cases where the correct answer is ranked within the top 3 predictions. The Hits@3 value is 0.26, indicating that the correct answer appears in the top 3 predictions in 26\% of the cases.
    \item \textbf{Hits@1}: It measures the proportion of test cases where the correct answer is ranked first among the predictions. The Hits@1 value is 0.21, indicating that the correct answer is ranked first in 21\% of the cases.
\end{itemize}

The values indicate an acceptable performance in predicting edges in the knowledge graph. However, it is important to note that the small size of the knowledge graph and its limited accuracy may have influenced the model's performance. The model's performance is expected to improve as the knowledge graph grows and its accuracy improves.

\subsubsection*{\textbf{The workflow}} \hfill

The list below outlines the logical workflow of our implemented solution:

\begin{enumerate}
  \item Registration of new dataset: A new dataset is registered in the database as part of the data ingestion process.
  \item Integration with domain knowledge graph: The extracted context information is incorporated into the domain knowledge graph. This graph contains a historical record of datasets, contexts, and corresponding assessment plans.
  \item Extracting triplet relations: Triplet relations with their respective weights are extracted from the knowledge graph. These include all the attributes of the new dataset, in addition to the quality rules and dimensions. The model is then trained for 5000 epochs to optimize the relationships.
  \item Link prediction: The final step involves predicting the links between each attribute of the new dataset and the corresponding quality rules and the quality rules with the related dimensions within the knowledge graph. Moreover, all predicted links include an injected numerical attribute presenting the weight. This prediction enables the creation of a comprehensive data quality plan for the new dataset.
\end{enumerate}

\subsubsection*{\textbf{Predicting a data quality assessment plan for radiation data}} \hfill

\begin{figure}[htbp]
\centering
\centerline{\includegraphics[width=\columnwidth]{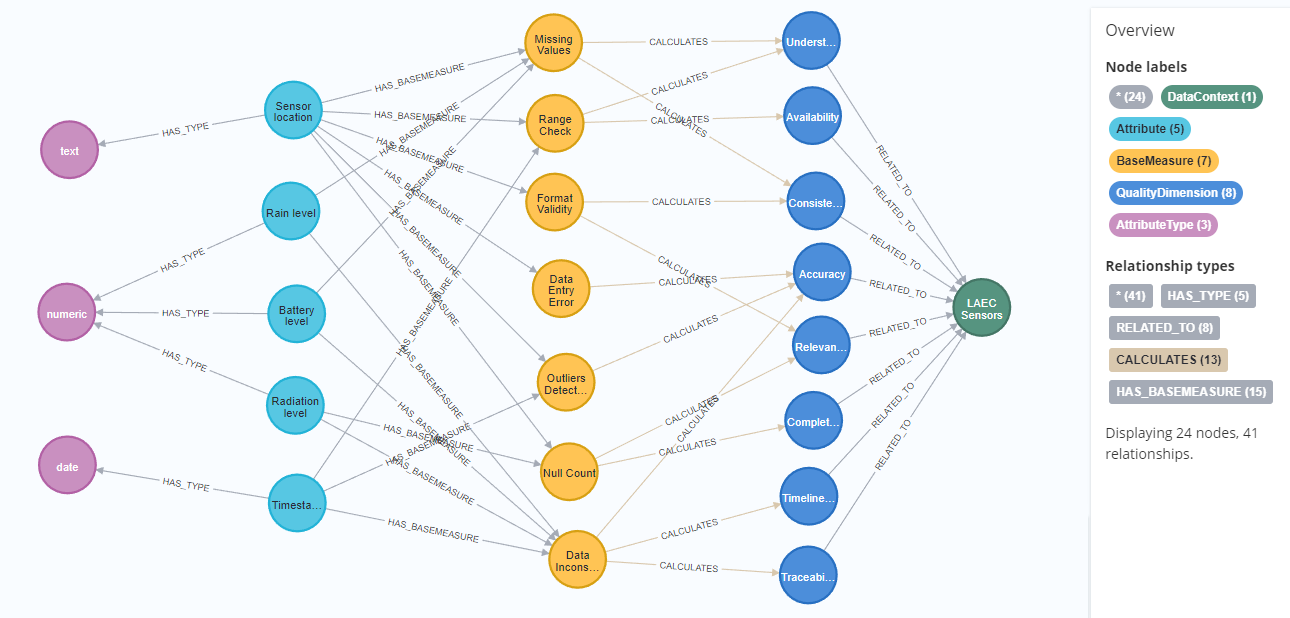}}
\caption{Retrieved data quality assessment plan using AmpliGraph}
\label{fig:KGEPlan}
\end{figure}

As a result, the generated data quality assessment plan encompassed seven quality rules and eight quality dimensions (Figure~\ref{fig:KGEPlan}). Unlike previous approaches, all attributes from the input dataset were linked to different quality rules, ensuring a complete quality assessment plan. Additionally, weights were assigned to all relevant edges.

\subsection{Comparison and Discussion of the results}

Our proposed approach offers several advantages compared to the initial BIGQA context analyzer component. When using the BIGQA context analyzer, we retrieved a data quality assessment plan from an existing radiation monitoring data context in the knowledge graph when applied to a real-world data set collected from radiation sensors. However, not all attributes were listed due to schema differences between the input data set and the matched plan, leading to an incomplete assessment. In contrast, we generated a comprehensive data quality assessment plan using this new approach that included all attributes, ensuring a more accurate evaluation. Additionally, automation is more easily achievable with this approach, as it does not rely on exact schema matching and reduces the need for user intervention. This eliminates the potential for user error and increases the efficiency of the quality assessment process, resulting in more reliable outcomes.

One possible limitation of our proposed solution is the potential loss of interpretability. Knowledge graph embedding models typically learn low-dimensional vector representations for entities and relations in the knowledge graph, making it challenging to understand the underlying logic and reasoning behind the generated embeddings. This lack of interpretability may limit the ability to explain the decision-making process and the specific factors contributing to the quality assessment plan. Additionally, both approaches may require substantial computational resources and time for training, especially when dealing with large and complex knowledge graphs.

As a result, the proposed solution utilizing knowledge graph embeddings should replace the initial BIGQA context analyzer component, as it offers significant advantages such as generating a comprehensive data quality assessment plan and enabling easier automation without relying on exact schema matching.

\section{Conclusion and Future Work}

Data quality assessment is crucial for ensuring the reliability of data-driven applications. However, manual data quality inspection is challenging and time-consuming, especially when dealing with big data. This research proposed a novel approach for automating data quality assessment. The limitations found in the literature were addressed by developing a new representation of the data context characteristics and injecting numeric edge attributes into the knowledge graph embedding model. The evaluation of the proposed solution using a real-world dataset from radiation sensors demonstrated its effectiveness in generating a more accurate data quality assessment plan than the previously implemented BIGQA context-aware context analyzer component.

The proposed solution overcame schema differences and incomplete attribute listing limitations, ensuring a comprehensive data quality assessment for diverse datasets. The automation capabilities of the new approach reduce the reliance on manual intervention, leading to more efficient and scalable data quality assessment processes.

However, using knowledge graph embedding models may introduce challenges related to interpretability, as the underlying logic and reasoning of the generated embeddings may need to be more easily understood. Furthermore, training knowledge graph embedding models can be computationally intensive, requiring adequate resources and time.

We plan to evaluate our solution using a larger and more precise knowledge graph in future work. Additionally, we will explore opportunities to enhance the interpretability of the knowledge graph embedding model and optimize the training process. Moreover, we aim to extend our proposed solution to handle different data context characteristics and incorporate additional quality dimensions beyond those studied in this research. Finally, we will investigate strategies to implement real-time machine-learning techniques to reduce computation time and enable real-time data quality assessment plan generation.

\bibliographystyle{unsrt}
\bibliography{Ref}

\end{document}